\pdfoutput=1

\documentclass[11pt]{article}

\usepackage{acl2021}

\usepackage{times}
\usepackage{latexsym}

\usepackage{graphicx}
\usepackage{float}
\usepackage[shortlabels]{enumitem}
\usepackage{mathtools}
\usepackage{tabularx, booktabs, subcaption, multirow}
\usepackage{adjustbox}
\newcommand{\ra}[1]{\renewcommand{\arraystretch}{#1}}
\usepackage{hyperref}

\usepackage{longtable, multicol}

\usepackage{algorithm}
\usepackage{algpseudocode}

\usepackage[
  separate-uncertainty = true,
  multi-part-units = repeat
]{siunitx}

\usepackage{array}
\newcolumntype{C}[1]{>{\centering\arraybackslash}m{#1}}
\newcolumntype{L}{>{\arraybackslash}m{2cm}}

\usepackage[T1]{fontenc}

\usepackage[utf8]{inputenc}

\usepackage{microtype}

\usepackage{soul}
\definecolor{MyGreen}{RGB}{144,238,144} 
\DeclareRobustCommand{\hlgreen}[1]{{\sethlcolor{YellowGreen}\hl{#1}}}

\title{Multi-Narrative Semantic Overlap Task: Evaluation and Benchmark}

\author{Naman Bansal, Mousumi Akter and Shubhra Kanti Karmaker Santu \\
        BDI Lab, Auburn University \\ 
        \{nzb0040, mza0170, sks0086\}@auburn.edu}

\begin{document}

\maketitle

\begin{abstract}
In this paper, we introduce an important yet relatively unexplored NLP task called \textit{Multi-Narrative Semantic Overlap} (MNSO), which entails generating a \textit{Semantic Overlap} of multiple alternate narratives. 
As no benchmark dataset is readily available for this task, we created one by crawling $2,925$ narrative pairs from the web and then, went through the tedious process of manually creating $411$ different ground-truth semantic overlaps by engaging human annotators.  
As a way to evaluate this novel task, we first conducted a systematic study by borrowing the popular \textit{ROUGE} metric from text-summarization literature and discovered that \textit{ROUGE} is not suitable for our task. 
Subsequently, we conducted further human annotations/validations to create $200$ document-level and $1,518$ sentence-level ground-truth labels which helped us formulate a new precision-recall style evaluation metric, called \textbf{SEM-F1} (semantic F1).
Experimental results show that the proposed SEM-F1 metric yields higher correlation with human judgement as well as higher inter-rater-agreement compared to ROUGE metric. 

\end{abstract}


\section{Introduction}
\label{sec:intro}



In this paper, we look deeper into the challenging yet relatively under-explored area of automated understanding of multiple alternative narratives. 
To be more specific, we formally introduce a new NLP task called \textit{Multi-Narrative Semantic Overlap} (MNSO) and conduct the first systematic study of this task by creating a benchmark dataset as well as proposing a suitable evaluation metric for the task.
MNSO essentially means the task of extracting / paraphrasing / summarizing the \textit{overlapping information} from multiple alternative narratives coming from disparate sources. 
In terms of computational goal, we study the following research question: 

\textit{Given two distinct narratives $N_1$ and $N_2$ of some event $e$ expressed in \textbf{unstructured natural language} format, how can we extract the overlapping information present in both $N_1$ and $N_2$? }



\begin{figure*}[t]
    \centering
    \includegraphics[width=0.85\textwidth]{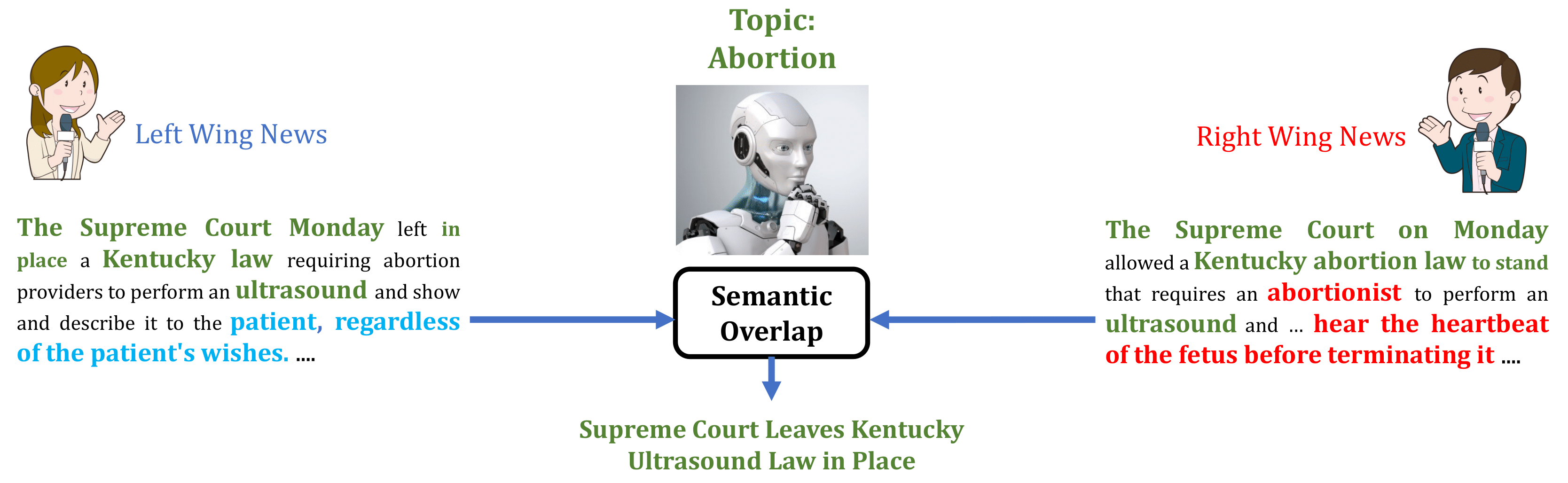}
    \vspace{-2mm}
    \caption{A toy use-case for Semantic Overlap Task (\textit{TextOverlap}). 
    A news on topic abortion has been presented by two news media (left-wing and right-wing). 
    ``Green'' Text denotes the overlapping information from both news media, while ``Blue'' and ``Red'' text denotes the respective biases of \textit{left} and \textit{right} wing. 
    A couple of real examples from the benchmark dataset are mentioned in the appendix.}
    \label{fig:semantic_intersection}
\end{figure*}


Figure~\ref{fig:semantic_intersection} shows a toy example of \textit{MNSO} task, where the \textit{TextOverlap}\footnote{We'll be using the terms \textit{TextOverlap} operator and \textit{Semantic Overlap} interchangeably throughout the paper.} ($\cap_{O}$) operation is being applied on two news articles. 
Both articles cover the same story related to the topic ``abortion'', however, they report from different political perspectives, i.e., one from \textit{left} wing and the other from \textit{right} wing. 
For greater visibility, ``Left'' and ``Right'' wing reporting biases are represented by \textit{blue} and \textit{red} text respectively. 
\textit{Green} text denotes the common information in both news articles. 
The goal of \textit{TextOverlap} ($\cap_{O}$) operation is to extract the overlapping information conveyed by the \textit{green} text.


At first glance, the MNSO task may appear similar to traditional multi-document summarization task where the goal is to provide an overall summary of the (multiple) input documents; however, the difference is that for \textit{MNSO}, the goal is to provide summarized content with an additional constraint, i.e., the commonality criteria. 
There is no current baseline method as well as existing dataset that exactly match our task; more importantly, it is unclear which one is the right evaluation metric to properly evaluate this task. 
As a starting point, we frame \textit{MNSO} as a constrained  seq-to-seq task where the goal is to generate a natural language output which conveys the overlapping information present in multiple input text documents. 
However, the bigger challenge we need to address first is the following: 1) How can we evaluate this task? and 2) How would one create a benchmark dataset for this task? 
To address these challenges, we make the following contributions in this paper.


\begin{enumerate}[
leftmargin=*,
]
    \item We formally introduce \textit{Multi-Narrative Semantic Overlap} (MNSO) as a new NLP task and conduct the first systematic study by formulating it as a constrained summarization problem. 
    
    \item We create and release the first benchmark data-set consisting of $2,925$ alternative narrative pairs for facilitating research on the \textit{MNSO} task.  
    Also, we went through the tedious process of manually creating $411$ different ground-truth semantic intersections and conducted further human annotations/validations to create $200$ document-level and $1,518$ sentence-level ground-truth labels to construct the dataset.
    
    \item As a starting point, we experiment with \textit{ROUGE}, a widely popular metric for evaluating text summarization tasks and demonstrate that \textit{ROUGE} is NOT suitable for evaluation of \textit{MNSO} task.
    
    \item We propose a new precision-recall style  evaluation metric, \textbf{SEM-F1} (semantic F1), for evaluating the \textit{MNSO} task. 
    Extensive experiments show that new SEM-F1 improves the inter-rater agreement compared to the traditional \textit{ROUGE} metric, and also, shows higher correlation with human judgments.
    
\end{enumerate}



\section{Related Works}
\label{sec:related_works}

The idea of semantic text overlap is not entirely new, \citep{karmaker2018sofsat} imagined a hypothetical framework for performing comparative text analysis, where, \textit{TextOverlap} was one of the ``hypothetical'' operators along with \textit{TextDifference}, but the technical details and exact implementation were left as a future work. 
In our work, we only focus on \textit{TextOverlap}.

As \textit{TextOverlap} can be viewed as a multi-document summarization task with additional commonality constraint, text summarization literature is the most relevant to our work.  
Over the years, many paradigms for document summarization have been explored \citep{zhong2019searching}. 
The two most popular among them are \textit{extractive} approaches \citep{cao-etal-2018-retrieve, narayan2018ranking, wu2018learning, zhong2020extractive} and \textit{abstractive} approaches \citep{bae2019summary, hsu2018unified, liu2017generative, nallapati2016abstractive}. 
Some researchers have also tried combining extractive and abstractive approaches \citep{chen2018fast, hsu2018unified, zhang2019pegasus}. 

Recently, encoder-decoder based neural models have become really popular for abstractive summarization \citep{rush2015neural, chopra2016abstractive, zhou2017selective, paulus2017deep}. 
It has become even prevalent to train a general language model on huge corpus of data and then transfer/fine-tune it for the summarization task \citep{radford2019better, devlin-etal-2019-bert, lewis2019bart, xiao2020ernie, yan2020prophetnet, zhang2019pegasus, raffel2019exploring}.
Summary length control for abstractive summarization has also been studied \citep{kikuchi2016controlling, fan2017controllable, liu2018controlling, fevry2018unsupervised, schumann2018unsupervised, makino2019global}. 
In general, multiple document summarization \citep{goldstein2000multi, yasunaga2017graph, zhao2020summpip, ma2020multi, meena2014survey} is more challenging than single document summarization. 
However, \textit{MNSO} task is different from traditional multi-document summarization tasks in that the goal here is to summarize content with an \textit{overlap} constraint, i.e., the output should only contain the common information from both input narratives.


Alternatively, one could aim to recover verb predicate-alignment structure \citep{roth2012aligning, xie2008event, wolfe2013parma} from a sentence and further, use this structure to compute the overlapping information \citep{wang2009recognizing, shibata2012predicate}.
Sentence Fusion is another related area which aims to combine the information from two given sentences with some additional constraints \citep{barzilay1999information, marsi2005explorations, krahmer2008query, thadani2011towards}. 
A related but simpler task is to retrieve parallel sentences \citep{cardon2019parallel, nie1999cross, murdock2005translation} without performing an actual intersection. 
However, these approaches are more targeted towards individual sentences and do not directly translate to arbitrarily long documents. 
Thus, MNSO task is still an open problem and there is no existing dataset, method or evaluation metric that have been systematically studied.



Along the evaluation dimension, \textit{ROUGE}~\cite{lin2004rouge} is perhaps the most commonly used metric today for evaluating automated summarization techniques; due to its simplicity and automation. 
However, \textit{ROUGE} has been criticized a lot for primarily relying on lexical overlap~\cite{DBLP:conf/interspeech/Nenkova06,DBLP:conf/naacl/ZhouLMH06,DBLP:conf/lrec/CohanG16} of n-grams. As of today, around $192$ variants of \textit{ROUGE} are available~\cite{DBLP:conf/emnlp/Graham15} including \textit{ROUGE} with word embedding~\cite{DBLP:conf/emnlp/NgA15} and synonym~\cite{DBLP:journals/corr/abs-1803-01937}, graph-based lexical measurement~\cite{DBLP:conf/emnlp/ShafieiBavaniEW18}, Vanilla \textit{ROUGE}~\cite{DBLP:conf/acl/YangLLLL18} and highlight-based \textit{ROUGE}~\cite{DBLP:conf/acl/HardyNV19}. 
However, there has been no study yet whether \textit{ROUGE} metric is appropriate for evaluating the \textit{Semantic Intersection} task, which is one of central goals of our work.



\section{Motivation and Applications}
\label{sec:motivation}

Multiple alternative narratives appear frequently across many domains like education, health, military, security and privacy etc (detailed use-cases for each domain are provided in appendix). 
Indeed, MNSO/TextOverlap operation can be very useful to digest such multi-narratives (from various perspectives) at scale and speed and, consequently, enhance the following important tasks as well.







\noindent \textbf{\underline{Information Retrieval/Search Engines}:} Given a query, one could summarize the common information (\textit{TextOverlap}) from the top $k$ documents fetched by a search engine and display it as additional information to the user. 

\noindent \textbf{\underline{Question Answering}:} Given a particular question, the system could aim to provide a more accurate answer based on multiple evidence from various source documents and generate the most common answer by applying \textit{TextOverlap}.

\noindent \textbf{\underline{Robust Translation}:} Suppose you have multiple translation models which translates a given document from language $A$ to language $B$. One could further apply the \textit{TextOverlap} operator on the translated documents and get a robust translation.

In general, MNSO task could be employed in any setting where we have comparative text analysis.

\section{Problem Formulation}
\label{sec:problem_formulation}
What is \textit{Semantic Overlap}? 
This is indeed a philosophical question and there is no single correct answer (various possible definitions are mentioned in appendix section \ref{app_sec:other_definitions}). 
To simplify notations, let us stick to having only two documents $D_A$ and $D_B$ as our input since it can easily be generalized in case of more documents using \textit{TextOverlap} repeatedly. 
Also, let us define the output as $D_{O} \gets D_A \cap_O D_B$. 
A human would mostly express the output in the form of natural language and this is why, we frame the \textit{MNSO} task as a constrained multi-seq-to-seq (text generation) task where the output text only contains information that is present in both the input documents.
We also argue that brevity (minimal repetition) is a desired property of \textit{Semantic Overlap} and thus, we frame \textit{MNSO} task as a constrained summarization problem to ensure brevity.
For example, if a particular piece of information or quote is repeated twice in both the documents, we don't necessarily want it to be present in target overlap summary two times.
The output can either be extractive summary or abstractive summary or a mixture of both, as per the use case. 
This task is inspired by the set-theoretic intersection operator. 
However, unlike set-intersection, our \textit{Text Overlap} does not have to be the \textit{maximal} set. 
The aim is summarize the overlapping information in an abstractive fashion. 
Additionally, \textit{Semantic Overlap} should follow the \textit{commutative} property i.e $D_A \cap_O D_B = D_B \cap_O D_A$.



\section{The Benchmark Dataset}
\label{sec:allsides_dataset}
As mentioned in section~\ref{sec:intro}, there is no existing data-set which we could readily use to evaluate the \textit{MNSO} task\footnote{Multi-document summarization datasets can not be utilized in this scenario as their reference summaries do not follow the semantic overlap constraint.}. 
To address this challenge, we crawled data from \href{https://www.allsides.com}{AllSides.com}.
AllSides is a third-party online news forum which exposes people to news and information from all sides of the political spectrum so that the general people can get an ``unbiased'' view of the world. 
To achieve this, AllSides displays each day’s top news stories from news media widely-known to be affiliated with different sides of the political spectrum including “Left” (e.g., New York Times, NBC News),
and “Right” (e.g., Townhall, Fox News) wing media.
AllSides also provides their own \textit{factual} description of the reading material, labeled as “Theme” so that readers can see the so-called ``neutral'' point-of-view. 
Table \ref{tab:allsides_dataset} gives an overview of the dataset created by crawling from AllSides.com, which consists of news articles (from at least one “Left” and one “Right” wing media) covering $2,925$ 
events in total and also having a minimum length of ``theme-description'' to be 15 words. 
Given two narratives (“Left” and “Right”), we used the theme-description as a proxy for ground-truth \textit{TextOverlap}. 
We divided this dataset into testing data (described next) and training data (remaining samples) and their statistics in provided in appendix (table \ref{tab:dataset_statistics}).

\begin{table}[!htb]
\footnotesize
\centering
\adjustbox{max width=\columnwidth}{%
\begin{tabular}{ll}
\toprule
\textbf{Feature} & \textbf{Description} \\
\midrule
theme                     & headlines by AllSides \\
theme-description         & news description by AllSides \\
right/left head    & right/left news headline \\
right/left context & right/left news description \\
\bottomrule
\end{tabular}
}
\caption{Overview of dataset scraped from AllSides}
\label{tab:allsides_dataset}
\vspace{-2mm}
\end{table}

\noindent{\bf Human Annotations\footnote{The dataset and manual annotations can be found in supplementary folder.}:} We decided to involve human volunteers to annotate our testing samples in order to create multiple human-written ground-truth semantic overlaps for each event narrative pairs. This helped in creating a comprehensive testing benchmark for more rigorous evaluation. 
Specifically, we randomly sampled $150$ narrative pairs (one from “Left” wing and one from “Right” wing) and then asked $3$ (three) humans to write a a natural language description which conveys the semantic overlap of the information present in both narratives describing each event.

After the first round of annotation, we immediately observed a discrepancy among the three annotators in terms of the \textit{real} definition of  ``semantic overlap''. For example, one annotator argued that \textit{Semantic Overlap} of two narratives is non-empty as long as there is an overlap along one of the \textit{5W1H} facets (Who,  What, When, Where, Why and How), while another annotator argued that overlap in only one facet is not enough to decide whether there is indeed a semantic overlap. As an example, one of the annotators wrote only ``Donald Trump'' as the \textit{Semantic Overlap} for a couple of cases where the narratives were substantially different, while others had those cases marked as ``empty set''. 

To mitigate this issue, we only retained the narrative-pairs where at least two of the annotators wrote minimum 15 words as their ground-truth semantic overlap, with the hope that a human written description will contain 15 words or more only in cases where there is indeed a ``significant'' overlap between the two original narratives. 
This filtering step gave us a test set with $137$ samples where each sample had 4 ground-truth semantic overlaps, \textit{one} from AllSides and \textit{three} from human annotators. 



\section{Evaluating MNSO Task using ROUGE}
\label{sec:case_study_rouge}
As \textit{ROUGE}~\cite{lin2004rouge} is the most popular metric used today for evaluating summarization techniques; we first conducted a case-study with \textit{ROUGE} as the evaluation metric for MNSO task. 

\subsection{Methods Used in the Case-Study}
\label{subsec:models}
We experimented with multiple SoTA pre-trained abstractive summarization
models as a proxy for \textit{Semantic-Overlap} generators. 
These models are: 1) \textbf{\href{https://huggingface.co/WikinewsSum/bart-large-cnn-multi-en-wiki-news}{BART}} \cite{lewis2019bart}, fine tuned on CNN and multi english Wiki news datasets,
2) \textbf{\href{https://huggingface.co/google/pegasus-cnn_dailymail}{Pegasus}} \cite{zhang2019pegasus}, fine tuned on CNN and Daily mail dataset, and
3) \textbf{\href{https://huggingface.co/WikinewsSum/t5-base-multi-en-wiki-news}{T5}} \cite {raffel2019exploring}, fine tuned on multi english Wiki news dataset. 
As our primary goal is to construct a benchmark data-set for the \textit{MNSO} task and establish an appropriate metric for evaluating this task, experimenting with only 3 abstractive summarization models is not a barrier to our work.
Proposing a custom method fine-tuned for the \textit{Semantic-Overlap} task is an orthogonal goal to this work and we leave it as a future work. 
Also, we'll use the phrases ``summary'' and ``overlap-summary'' interchangeably from here. 
To generate the summary, we concatenate a narrative pair and feed it directly to the model.




For evaluation, we first evaluated the machine generated overlap summaries for the 137 manually annotated testing samples using the ROUGE metric \citep{lin2004rouge} and followed the procedure mentioned in the paper to compute the ROUGE-$F_1$ scores with multiple reference summaries.
More precisely, since we have 4 reference summaries, we got 4 precision, recall pairs which are used to compute the corresponding $F_1$ scores. 
For each sample, we took the max of these  $4 F_1$ scores and averaged them out across the test dataset. 
The ROUGE scores can be seen in the table \ref{tab:rouge_score} in appendix.

\begin{table}[!t]
\centering
\begin{adjustbox}{width=\linewidth,center}
\begin{tabular}{cccccccccccc}
\toprule
\multicolumn{12}{c}{\textbf{Pearson's Correlation Coefficients}} \\ \cmidrule{1-12}
& \multicolumn{3}{c}{\textbf{R1}} &  
& \multicolumn{3}{c}{\textbf{R2}} &  
& \multicolumn{3}{c}{\textbf{RL}} \\ 
\midrule
 & I\textsubscript{1} & I\textsubscript{2} & I\textsubscript{3} &&
I\textsubscript{1} & I\textsubscript{2} & I\textsubscript{3} &&
I\textsubscript{1} & I\textsubscript{2} & I\textsubscript{3}  \\
\cmidrule{1-4} \cmidrule{5-8} \cmidrule{9-12}
I\textsubscript{2} & $\mathbf{0.62}$   & ---     &   && $\mathbf{0.65}$   & ---     &   && $\mathbf{0.69}$   & ---     &   \\ 
I\textsubscript{3} & $\mathbf{0.3}$  & $\mathbf{0.38}$   & ---     && $\mathbf{0.27}$  & $\mathbf{0.37}$  & --- && $\mathbf{0.27}$  & $\mathbf{0.44}$  & ---    \\ 
I\textsubscript{4} & $\mathbf{0.17}$   & $\mathbf{0.34}$  & $\mathbf{0.34}$    && $0.14$  & $\mathbf{0.33}$   & $\mathbf{0.21}$   && $\mathbf{0.18}$   & $\mathbf{0.35}$  & $\mathbf{0.33}$  \\
\cmidrule{1-12}
\textbf{Average} & 
\multicolumn{3}{c}{\textbf{0.36}} &&
\multicolumn{3}{c}{\textbf{0.33}} &&
\multicolumn{3}{c}{\textbf{0.38}} \\


\bottomrule
\end{tabular}
\end{adjustbox}
\caption{Max (across 3 models) Pearson's correlation between the $F_1$ ROUGE scores corresponding to different annotators. 
Here I\textsubscript{i} refers to the $i^{th}$ annotator where $i \in \{ 1, 2, 3, 4 \}$ and ``Average'' row represents average correlation of the max values across annotators. 
Boldface values are statistically significant at p-value $<0.05$. 
For $5$ out of $6$ annotator pairs, the correlation values are quite small ($\leq 0.50$), thus, implying the poor inter-rated agreement with regards to ROUGE metric.}
\vspace{-2mm}
\label{tab:inter_annotator_agreement_rouge}
\end{table} 
\subsection{Results and Findings}
We computed Pearson's correlation coefficients between each pair of ROUGE-$F_1$ scores obtained using all of the $4$ reference overlap-summaries ($3$ human written summary and $1$ AllSides theme description) to test the robustness of \textit{ROUGE} metric for evaluating the \textit{MNSO} task. 
The corresponding correlations are shown in table \ref{tab:inter_annotator_agreement_rouge}. 
For each annotator pair, we report the maximum (across 3 models) correlation value.
The average correlation value across annotators is $0.36$, $0.33$ and $0.38$ for R1, R2 and RL respectively;
suggesting that ROUGE metric is not stable across multiple human-written overlap-summaries and thus, \textit{unreliable}. 
Indeed, only one out the $6$ different annotator pairs has a value greater than $0.50$ for all the $3$ ROUGE metrics (R1, R2, RL), which is \textit{problematic}.




\section{Can We Do Better than ROUGE?}
\label{sec:summary_label_annotations}
Section \ref{sec:case_study_rouge} shows that ROUGE metric is unstable across multiple reference overlap-summaries. 
Therefore, an immediate question is: Can we come up with a better metric than ROUGE?
To investigate this question, we started by manually assessing the machine-generated overlap summaries to check whether humans agree among themselves or not.  

\subsection{Different trials of Human Judgement}
\label{subsec:direct_annotation}

\textbf{Assigning a Single Numeric Score:} As an initial trial, we decided to first label 25 testing samples using two human annotators (we call them label annotators $L_1$ and $L_2$). 
Both label-annotators read each of the $25$ narrative pairs as well as the corresponding system generated overlap-summary (generated by fine-tuned BART)
and assigned a numeric score between 1-10 (inclusive).
%
This number reflects their judgement/confidence about how accurately the system-generated summary captures the \textit{actual} overlap of the two input narratives. 
\textit{Note that, the reference overlap summaries were not included in this label annotation process and the label-annotators judged the system-generated summary exclusively with respect to the input narratives}. 
%
To quantify the agreement between human scores, we computed the Kendall rank correlation coefficient (or Kendall's Tau) between two annotator labels since these are ordinal values. 
However, to our disappointment, the correlation value was $0.20$ with p-value being $0.22$\footnote{The higher p-value means that the correlation value is insignificant because of the small number of samples, but the aim is to first find a labelling criterion where human can agree among themselves.}. 
This shows that even human annotators are disagreeing among themselves and we need to come up with a better labelling guideline to reach a reasonable agreement among the human annotators. 
 
On further discussions among annotators, we realized that one annotator only focused on \textit{preciseness} of the intersection summaries, whereas the other annotator took both \textit{precision} and \textit{recall} into consideration. 
Thus, we decided to next assign two  separate scores for precision and recall.

\begin{table}[!t]\scriptsize 
\centering

\begin{tabular}{cccccc}
\toprule
\multicolumn{6}{c}{\textbf{Human agreement in terms of Kendall Tau}} \\
\cmidrule{2-6}
& \multicolumn{2}{c}{\textbf{Precision}} && \multicolumn{2}{c}{\textbf{Recall}} \\ \midrule
& L\textsubscript{1} & L\textsubscript{2} && L\textsubscript{1} & L\textsubscript{2} \\
\cmidrule{2-3} \cmidrule{5-6} 
L\textsubscript{2} & $0.52$ & --- && $0.37$ & ---\\
L\textsubscript{3} & $0.18$ & $0.29$ && $0.31$ & $0.54$\\ 

\cmidrule{1-6}
\textbf{Average} & 
\multicolumn{2}{c}{$\mathbf{0.33}$} &&
\multicolumn{2}{c}{$\mathbf{0.41}$} \\

\bottomrule
\end{tabular}

\caption{Kendall's rank correlation coefficients among the precision and recall scores for pairs of human annotators ($25$ samples). L\textsubscript{i} refers to the $i^{th}$ label annotator.}
\label{tab:two-number-agreeement}
\vspace{-4mm}
\end{table}
\noindent\textbf{Precision-Recall Inspired Double Scoring:} 
This time, three label-annotators ($L_1$, $L_2$ and $L_3$) assigned two numeric scores between 1-10 (inclusive) for the same set of $25$ system generated summaries. 
These numbers represented their belief about how precise the system-generated summaries were (the precision score) and how much of the actual ground-truth overlap-information was covered by the same (the recall score).
\textit{Also note that, labels were assigned exclusively with respect to the input narratives only}. 
As the assigned numbers represent ordinal values (i.e. can't be used to compute $F_1$ score), we compute the Kendall's rank correlation coefficient among the precision scores and recall scores of all the annotator pairs separately. 
The corresponding correlation values can be seen in the table \ref{tab:two-number-agreeement}. 
As we notice, there is definitely some improvement in agreement among annotators compared to the one number annotation in \ref{subsec:direct_annotation}.
However, the average correlation is still $0.33$ and $0.41$ for precision and recall respectively, much lower than the $0.5$. 




\subsection{Sentence-wise Scoring}
\label{subsec:sentence_wise_annotation}
From the previous trials, we realised the downsides of assigning one/two numeric scores to judge an entire system-generated overlap-summary. 
Therefore, as a next step, we decided to assign overlap labels to the each sentence within the system-generated overlap summary and use those labels to compute an overall precision and recall score.

\begin{table}[!t]\scriptsize
\centering

\begin{tabular}{cccccc}
\toprule
\multicolumn{6}{c}{\textbf{Human agreement in terms of Kendall's Rank Correlation}} \\
\cmidrule{2-6}
& \multicolumn{2}{c}{\textbf{Precision}} && \multicolumn{2}{c}{\textbf{Recall}} \\ \midrule
& L\textsubscript{1} & L\textsubscript{2} && L\textsubscript{1} & L\textsubscript{2} \\
\cmidrule{2-3} \cmidrule{5-6} 
L\textsubscript{2} & $0.68$ & --- && $0.75$ & ---\\
L\textsubscript{3} & $0.59$ & $0.64$ && $0.69$ & $0.71$\\ 

\cmidrule{1-6}
\textbf{Average} & 
\multicolumn{2}{c}{$\mathbf{0.64}$} &&
\multicolumn{2}{c}{$\mathbf{0.72}$} \\

\bottomrule
\end{tabular}

\caption{Average precision and recall Kendall rank correlation coefficients between sentence-wise annotation for different annotators. L\textsubscript{i} refers to the $i^{th}$ label annotator.
All values are statistically significant (p<0.05).}
\label{tab:kendall-sentence-wise-annotation-agreeement}
\vspace{-4mm}
\end{table}
\noindent\textbf{Overlap Labels}: Label-annotators ($L_1$, $L_2$ and $L_3$) were asked to look at a machine-generated sentence and determine if the core information conveyed by it is either absent, partially present or present in any of the four reference summaries (provided by ($I_1$, $I_2$, $I_3$ and $I_4$) and respectively, assign the label \textit{A}, \textit{PP} or \textit{P}. 
More precisely, if the human feels there is more than $75\%$ overlap (between each system-generated sentence and reference-summary sentence), assign label \textit{P}, else if the human feels there is less than $25\%$ overlap, assign label \textit{A}, and else, assign \textit{PP} otherwise. 
This sentence-wise labelling was done for 50 different samples (with $506$ sentences in total for system and reference summary), which resulted in total $3\times 506=1,518$ sentence-level ground-truth labels.

To create the overlap labels from precision perspective as described above, we concatenated all the 4 reference summaries to make one big reference summary and asked label-annotators ($L_1$, $L_2$ and $L_3$) to use it as a reference for assigning the overlap labels to each sentence within machine generated summary. 
We argue that if the system could generate a sentence conveying information which is present in any of the references, it should be considered a hit. 
For recall, label-annotators were asked to assign labels to each sentences in each of the 4 reference summaries separately (provided by ($I_1$, $I_2$, $I_3$ and $I_4$)), with respect to the machine generated summary.


\noindent\textbf{Inter-Rater-Agreement}: 
We use the Kendall rank correlation coefficient to compute the agreement among the ordinal labels assigned by human label annotators. 
Since there can be multiple sentences in the system generated or the reference summary, we flatten out the sentence labels and concatenate them for the entire dataset. 
To compute the Kendall Tau, we map the ordinal labels to numerical values using the mapping: $\{P:1, PP:0.5, A:0\}$. 
Table \ref{tab:kendall-sentence-wise-annotation-agreeement} shows that inter-annotator correlation for both precision and recall are $\geq 0.50$ and thus, signifying higher agreement among label annotators. 

\begin{table}[!t]\scriptsize
\centering
\begin{tabular}{c|c|ccc}
\toprule

\multicolumn{2}{c|}{\textbf{Label from Annotator B}} &  P   & PP  & A \\\cmidrule{1-5}
\multirow{3}{*}{\parbox{1.7cm}{\textbf{Label from Annotator A}}} & P                        & 1   & 0.5 & 0 \\ 
& PP                       & 0.5 & 1   & 0 \\ 
& A                        & 0   & 0   & 1 \\ 
\bottomrule
\end{tabular}
\caption{Reward function used to evaluate the labels assigned by two label annotators (or labels inferred using SEM-F1 metric and human annotated labels) for a given sentence (association between annotator pairs).}
\label{tab:reward_table}
\vspace{-2mm}
\end{table}


\noindent\textbf{Reward-based Inter-Rater-Agreement}: 
Alternatively, 
we first define a reward matrix (Table \ref{tab:reward_table}) which is used to compare the label of one annotator (say annotator A) against the label of another annotator (say annotator B) for a given sentence.
This reward matrix acts as a form of correlation between two annotators. 
Once reward has been computed for each sentence, one can compute the average precision and recall rewards for a given sample and accordingly, for the entire test dataset.
The corresponding reward scores can be seen in table \ref{tab:sentence-wise-annotation-agreeement}. 
Both precision and recall reward scores are high ($\geq 0.70$) for all the different annotator pairs, thus signifying, high inter label-annotator agreement. 

We believe, one of the reasons for higher reward/Kendall scores could be that sentence-wise labelling puts less cognitive load on human mind in contrast to the single or double score(s) for the entire overlap summary and accordingly, shows high agreement in terms of human interpretation. 
Similar observation is also noted in \citet{harman-over-2004-effects}. 

\begin{table}[!t]\small
\centering
\adjustbox{max width=\columnwidth}{%
\begin{tabular}{cccccc}
\toprule
\multicolumn{6}{c}{\textbf{Human agreement in terms of Reward function}} \\
\cmidrule{2-6}
& \multicolumn{2}{c}{\textbf{Precision}} && \multicolumn{2}{c}{\textbf{Recall}} \\ \midrule
& L\textsubscript{1} & L\textsubscript{2} && L\textsubscript{1} & L\textsubscript{2} \\
\cmidrule{2-3} \cmidrule{5-6} 
L\textsubscript{2} & $\SI{0.81 \pm 0.26}{}$ & --- && $\SI{0.85 \pm 0.11}{}$ & ---\\
L\textsubscript{3} & $\SI{0.79 \pm 0.26}{}$ & $\SI{0.7 \pm 0.31}{}$ && $\SI{0.8 \pm 0.16}{}$ & $\SI{0.77 \pm 0.17}{}$\\ 

\cmidrule{1-6}
\textbf{Average} & 
\multicolumn{2}{c}{$\mathbf{0.77}$} &&
\multicolumn{2}{c}{$\mathbf{0.81}$} \\
\bottomrule
\end{tabular}
}
\vspace{-2mm}
\caption{Average precision and recall reward scores (mean $\pm$ std) between sentence-wise annotation for different annotators. L\textsubscript{i} refers to the $i^{th}$ label-annotator.}
\label{tab:sentence-wise-annotation-agreeement}
\end{table}


\section{Semantic-F1: The New Metric}
\label{subsec:automatic_evaluation}
Human evaluation is costly and time-consuming.
Thus, one needs an automatic evaluation metric for large-scale experiments. 
But, how can we devise an automated metric to perform the sentence-wise precision-recall style evaluation discussed in the previous section? 
To achieve this, we propose a new evaluation metric called \textbf{SEM-F1}. 
The details of our \textbf{SEM-F1} metric are described in algorithm \ref{algo:automatic_evaluation} and the respective notations are mentioned in table \ref{tab:symbol_table}. 
$F_1$ scores are computed by the harmonic mean of the precision ($pV$) and recall ($rV$) values.
Algorithm \ref{algo:automatic_evaluation} assumes only one reference summary but can be trivially extended for multiple references. 
As mentioned previously, in case of multiple references, we concatenate them for precision score computation. 
Recall scores are computed individually for each reference summary and later, an average recall is computed across references. 

The basic intuition behind \textbf{SEM-F1} is to compute the sentence-wise similarity (e.g., cosine similarity using a sentence embedding model) to infer the semantic overlap/intersection between two sentences from both precision and recall perspective and then, combine them into $F_1$ score. 


\begin{table}[!htb]\footnotesize
\centering
\begin{tabularx}{\columnwidth}{lX}
\toprule
Notations & Description \\ 
\midrule
$S_G$ & Machines generated summary \\ 
$S_R$ & Reference summary \\ 
$T \coloneqq (t_l, t_u)$ & Tuple representing the lower and upper threshold values (between 0 and 1). \\ 
$M_E$ & Sentence embedding model \\
$pV, rV$ & Precision, Recall value for $(S_G, S_R)$ pair \\
\bottomrule

\end{tabularx}
\vspace{-2mm}
\caption{Table of notations for algorithm \ref{algo:automatic_evaluation}}
\label{tab:symbol_table}
\vspace{-5mm}
\end{table}


\setlength{\textfloatsep}{8pt}
\begin{algorithm}[!t]\footnotesize 
\caption{Semantic-F1 Metric}
\label{algo:automatic_evaluation}

\begin{algorithmic}[1]

    \State{\textbf{Given} $S_G, S_R, M_E$ }

    \State{$raw_{pV}, raw_{rV} \gets \textsc{CosineSim}(S_G, S_R, M_E)$}
    \Comment{Sentence-wise precision and recall values}
    
    \State{$pV \gets \textsc{Mean}(raw_{pV})$}
    \State{$rV \gets \textsc{Mean}(raw_{rV})$}
    \State{$f_1 \gets \dfrac{2*pV*rV}{pV+rV}$}
    
    
    \State{\textbf{return} $(f_1, pV, rV)$ }
    
\end{algorithmic}



\hrulefill
\begin{algorithmic}[1]
    \Procedure{CosineSim}{$S_G, S_R, M_E$}
        \State{$l_G \gets${No. of sentences in $S_G$}}
        \State{$l_R \gets${No. of sentences in $S_R$}}
        \State{\textbf{init}: $cosSs \gets zeros[l_G, l_R]$; $i \gets 0$}
    
        \For{each sentence $sG$ in $S_G$}
            \State{$E_{sG} \gets M_E(sG)$;$j \gets 0$} 
            \For{each sentence $sR$ in $S_R$}
                \State{$E_{sR} \gets M_E(sR)$} 
                \State{$cosSs[i, j] \gets Cos(E_{sG}, E_{sR})$}
            \EndFor
        \EndFor
        \State{$\boldsymbol{x}\gets$  Row-wise-max($cosSs$)}
        \State{$\boldsymbol{y}\gets$  Column-wise-max($cosSs$)}
        \State{\textbf{return} $(\boldsymbol{x}, \boldsymbol{y})$}
    \EndProcedure
\end{algorithmic}

        
        
        

\end{algorithm}


\begin{table*}[!htb]\footnotesize
\centering
\adjustbox{max width=\textwidth}{%
\begin{tabular}{p{1.9cm}lccccccc}
\toprule
\multicolumn{2}{c}{\multirow{2}{*}{\textbf{Reward/Kendall}}} &
\multicolumn{7}{c}{\textbf{Machine-Human Agreement in terms of Reward Function}}  \\
\cmidrule{3-9}
\multicolumn{2}{c}{} &
\multicolumn{1}{c}{$\mathbf{T=(25, 75)}$} & \multicolumn{1}{c}{$\mathbf{T=(35, 65)}$} & \multicolumn{1}{c}{$\mathbf{T=(45, 75)}$}& \multicolumn{1}{c}{$\mathbf{T=(55, 65)}$} & \multicolumn{1}{c}{$\mathbf{T=(55, 75)}$} & \multicolumn{1}{c}{$\mathbf{T=(55, 80)}$} & \multicolumn{1}{c}{$\mathbf{T=(60, 80)}$}\\ 
\midrule

\multirow{2}{*}{\parbox{1.9cm}{\textbf{Embedding: P-v1}}} 
& \textbf{Precision} & $0.75 / 0.57$ & $0.8/0.63$ & $0.76/0.59$ & $0.8/0.63$ & $0.78/0.6$ & $0.74/0.6$ & $0.73/0.58$ \\
\cmidrule{2-9}
& \textbf{Recall} & $0.66/0.54$ & $0.76/0.64$ & $0.73/0.66$ & $0.72/0.64$ & $0.69/0.63$ & $0.65/0.64$ & $0.61/0.6$ \\
\midrule
\multirow{2}{*}{\parbox{1.9cm}{\textbf{Embedding: STSB}}} 
& \textbf{Precision} & $0.73/0.6$ & $0.73/0.62$ & $0.73/0.6$ & $0.73/0.62$ & $0.73/0.63$ & $0.73/0.59$ & $0.73/0.58$ \\
\cmidrule{2-9}
& \textbf{Recall} & $0.63/0.55$ & $0.64/0.63$ & $0.63/0.6$ & $0.65/0.61$ & $0.65/0.61$ & $0.63/0.61$ & $0.64/0.59$ \\
\midrule
\multirow{2}{*}{\parbox{1.9cm}{\textbf{Embedding: USE}}} 
& \textbf{Precision} & $0.76/0.64$ & $0.76/0.66$ & $0.78/0.64$ & $0.78/0.64$ & $0.79/0.63$ & $0.78/0.62$ & $0.79/0.65$ \\
\cmidrule{2-9}
& \textbf{Recall} & $0.63/0.53$ & $0.66/0.6$ & $0.67/0.58$ & $0.68/0.61$ & $0.67/0.62$ & $0.64/0.62$ & $0.65/0.61$ \\

\bottomrule
\end{tabular}
}
\vspace{-2mm}
\caption{
Average Precision and Recall correlation (Reward score/Kendall correlation) between label-annotators (L\textsubscript{i}) and automatically inferred labels using SEM-F1 (average of 3 label annotators). 
The raw numbers for each annotators can be found in appendix (table \ref{tab:raw_average_rewards}). 
The results are shown for different embedding models (\ref{subsec:sem_f1_evaluation}) and multiple threshold levels $T=(t_l, t_u)$. 
Moreover, the both the Reward and Kendall values are consistent/stable across all the $5$ embedding models and threshold values.
}
\label{tab:average_rewards}
\vspace{-1mm}
\end{table*}

\subsection{Is SEM-F1 Reliable?}
\label{subsec:sem_f1_evaluation} 

The SEM-F1 metric computes cosine similarity scores between sentence-pairs from both precision and recall perspectives.
To see whether SEM-F1 metric correlates with human-judgement, we further converted the sentence-wise raw cosine scores into \textit{Presence} (P), \textit{Partial Presence} (PP) and \textit{Absence} (A) labels using some user-defined thresholds as described in algorithm \ref{algo:threshold}. 
This helped us to directly compare the SEM-F1 inferred labels against the human annotated labels.

As mentioned in section \ref{subsec:automatic_evaluation}, we utilized state-of-the-art sentence embedding models to encode sentences from both the model generated summaries and the human written narrative intersections.
To be more specific, we experimented with $3$ sentence embedding models: 
Paraphrase-distilroberta-base-v1 (\textit{P-v1}) \citep{reimers-2019-sentence-bert}, stsb-roberta-large (\textit{STSB}) \citep{reimers-2019-sentence-bert} and universal-sentence-encoder (\textit{USE}) \citep{cer2018universal}.
Along with the various embedding models, we also experimented with multiple threshold values used to predict the sentence-wise \textit{presence (P)}, \textit{partial presence (PP)} and \textit{absence (A)} labels to report the sensitivity of the metric with respect to different thresholds. 
These thresholds are: $(25, 75)$, $(35, 65)$, $(45, 75)$, $(55, 65)$, $(55, 75)$, $(55, 80)$, $(60, 80)$. 
For example, threshold range $(45, 75)$ means that if similarity score $<$ 45\%, infer label "absent", else if similarity score $\geq$ 75\%, infer label "present" and else, infer label ``partial-present''. 
Next, we computed the average precision and recall rewards for 50 samples annotated by label-annotators (L\textsubscript{i}) and the labels inferred by SEM-F1 metric. 
For this, we repeat the procedure of Table~\ref{tab:sentence-wise-annotation-agreeement}, but this time comparing human labels against `SEM-F1 labels'. 
The corresponding results are shown in Table~\ref{tab:average_rewards}.  
As we can notice, 
the average reward values are consistently high ($\geq 0.50$) for all the $3$ label-annotators (L\textsubscript{i}).  
Moreover, the reward values are consistent/stable across all the $3$ embedding models and threshold values, signifying 
that SEM-F1 is indeed robust 
across various sentence embeddings and threshold used. 

Following the procedure in table \ref{tab:kendall-sentence-wise-annotation-agreeement}, we also  compute the Kendall's Tau between human label annotators and automatically inferred labels using SEM-F1. 
Our results in table \ref{tab:average_rewards} are consistent with reward-based inter-rater-agreement and the correlation values are $\geq 0.50$ with little variation along various thresholds for both precision and recall. 


\begin{table}[!t]
\centering
\adjustbox{max width=\linewidth}{%
\begin{tabular}{cccccccccccc}
\toprule
& \multicolumn{3}{c}{\textbf{Random Annotation}} & &
\multicolumn{3}{c}{\textbf{Random Intersection}} & &
\multicolumn{3}{c}{\textbf{SEM-F1 Scores}}\\
& \multicolumn{3}{c}{\textbf{SEM-F1 Scores}} & &
\multicolumn{3}{c}{\textbf{SEM-F1 Scores}} & &
\multicolumn{3}{c}{\textbf{}}\\
\midrule
& \textbf{P-V1} &  \textbf{STSB} & \textbf{USE} & & 
\textbf{P-V1} &  \textbf{STSB} & \textbf{USE} & & 
\textbf{P-V1} &  \textbf{STSB} & \textbf{USE} \\

\cmidrule{2-4} \cmidrule{6-8} \cmidrule{10-12}
\textbf{BART} & $0.16$ & $0.21$ & $0.22$ & &
$0.21$ & $0.27$ & $0.27$ & & 
$0.65$ & $0.67$ & $0.67$ \\

\textbf{T5} & $0.17$ & $0.21$ & $0.23$ & &
$0.20$ & $0.26$ & $0.26$ & &
$0.58$ & $0.60$ & $0.60$ \\

\textbf{Pegasus} & $0.15$ & $0.20$ & $0.22$ & &
$0.19$ & $0.26$ & $0.26$ & &
$0.59$ & $0.60$ & $0.62$ \\

\cmidrule{1-12}
\textbf{Average} & $0.16$ & $0.21$ & $0.22$ & &
$0.20$ & $0.26$ & $0.26$ & &
$0.61$ & $0.62$ & $0.63$ \\
\bottomrule
\end{tabular}
}
\vspace{-2mm}
\caption{SEM-F1 Scores}
\label{tab:sem_f1_scores}
\end{table}
\subsection{SEM-F1 Scores for Random Baselines}
\label{sec: baseline_and_sem_f1_scores}
Here, we present the actual SEM-F1 scores for the three models described in section~\ref{subsec:models} along with scores for two intuitive baselines, namely, 1) Random Overlap 2) Random Annotation.

\noindent\textbf{Random Overlap:} For a given sample and model, we select a random overlap summary generated by the model out of the other 136 test samples. These random overlaps are then evaluated against 4 reference summaries using SEM-F1. 

\noindent\textbf{Random Annotation:} For a given sample, we select a random reference summary out of the other 4 references among the other 136 test samples. 
The model generated summaries are then compared against these Random Annotations/References to compute SEM-F1 scores as reported in table \ref{tab:sem_f1_scores}.

As we notice, there is approximately 40-45 percent improvement over the baseline scores suggesting SEM-F1 can indeed distinguish \textit{good} from \textit{bad}. 


\subsection{Pearson Correlation for SEM-F1}
Following the case-study based on ROUGE in section \ref{sec:case_study_rouge}, we again compute the Pearson's correlation coefficients between each pair of raw SEM-F1 scores obtained using all of the $4$ reference intersection-summaries.
The corresponding correlations are shown in table \ref{tab:inter_annotator_agreement_sem-f1}. 
For each annotator pair, we report the maximum (across 3 models) correlation value.
The average correlation value across annotators is $0.49$, $0.49$ and $0.54$ for \textbf{P-V1}, \textbf{STSB}, \textbf{USE} embeddings, respectively. 
This shows a clear improvement over the ROUGE metric suggesting that SEM-F1 is more accurate than ROUGE metric.

\begin{table}[!t]\footnotesize
\centering
\begin{adjustbox}{width=\linewidth,center}
\begin{tabular}{cccccccccccc}
\toprule
\multicolumn{12}{c}{\textbf{Pearson's Correlation Coefficients}} \\ \cmidrule{1-12}
& \multicolumn{3}{c}{\textbf{P-V1}} &  
& \multicolumn{3}{c}{\textbf{STSB}} &  
& \multicolumn{3}{c}{\textbf{USE}} \\ 
\midrule
 & I\textsubscript{1} & I\textsubscript{2} & I\textsubscript{3} &&
I\textsubscript{1} & I\textsubscript{2} & I\textsubscript{3} &&
I\textsubscript{1} & I\textsubscript{2} & I\textsubscript{3}  \\
\cmidrule{1-4} \cmidrule{5-8} \cmidrule{9-12}
I\textsubscript{2} & $\mathbf{0.69}$   & ---     &   && $\mathbf{0.65}$   & ---     &   && $\mathbf{0.71}$   & ---     &   \\ 
I\textsubscript{3} & $\mathbf{0.40}$  & $\mathbf{0.50}$   & ---     && $\mathbf{0.50}$  & $\mathbf{0.52}$  & --- && $\mathbf{0.51}$  & $\mathbf{0.54}$  & ---    \\ 
I\textsubscript{4} & $\mathbf{0.33}$   & $\mathbf{0.44}$  & $\mathbf{0.60}$    && $\mathbf{0.33}$  & $\mathbf{0.36}$   & $\mathbf{0.56}$   && $\mathbf{0.37}$   & $\mathbf{0.42}$  & $\mathbf{0.66}$  \\
\cmidrule{1-12}
\textbf{Average} & 
\multicolumn{3}{c}{\textbf{0.49}} &&
\multicolumn{3}{c}{\textbf{0.49}} &&
\multicolumn{3}{c}{\textbf{0.54}} \\
\bottomrule
\end{tabular}
\end{adjustbox}
\vspace{-2mm}
\caption{Max (across 3 models) Pearson's correlation between the SEM-F1 scores corresponding to different annotators. 
Here I\textsubscript{i} refers to the $i^{th}$ annotator where $i \in \{ 1, 2, 3, 4 \}$ and ``Average'' row represents average correlation of the max values across annotators.
All values are statistically significant at p-value $<0.05$.
}
\label{tab:inter_annotator_agreement_sem-f1}
\end{table} 


\section{Conclusions}
\label{conclusions}
In this work, we proposed a new NLP task, called Multi-Narrative Semantic Overlap (\textit{MNSO}) and created a benchmark dataset through meticulous human effort to initiate a new research direction.
As a starting point, we framed the problem as a constrained summarization task and showed that \textit{ROUGE} is not a reliable evaluation metric for this task. 
We further proposed a more accurate metric, called \textit{SEM-F1}, for evaluating \textit{MNSO} task.
Experiments show that SEM-F1 is more robust and yield higher agreement with human judgement.




\bibliography{main}
\bibliographystyle{acl_natbib}

\clearpage
\appendix
\section{Other definitions of Text Overlap}
\label{app_sec:other_definitions}

Below, we present a set of possible definitions of \textit{Semantic Overlap} to encourage the readers to think more about other alternative definitions. 

\begin{enumerate}[leftmargin=*,
]
    \item On a very simplistic level, one can think of \textit{Semantic Overlap} to be just the common words between the two input documents. 
    One can also include their frequencies of occurrences in such representation. 
    More specifically, we can define 
    $D_{ovlp}$ as a set of unordered pairs of words $w_i$ and their frequencies of common occurrences $f_i$, i.e., $D_{ovlp}=\{( w_i, f_i)\}$. 
    We can further extend this approach such that \textit{Semantic Overlap} is a set of common n-grams among the input documents. 
    More specifically, $D_{ovlp} = \{ \big((w_1, w_2, ..., w_n)_i, f_i \big)\}$ such that the n-grams, $(w_1, w_2, ..., w_n)_i$, is present in both $D_A$ (with frequency $f_{iA}$) and $D_B$ (with frequency $f_{iB}$) and $f_i = min(f_{iA}, f_{iB})$. 
    
    
    \item Another way to think of \textit{Semantic Overlap} is to find the common topics among two documents just like finding common object labels among two images \citep{alfassy2019laso}, by computing the joint probability of their topic distributions. 
    More specifically, \textit{Semantic Overlap} can be defined by the following joint probability distribution: $P(T_i | D_{ovlp}) = P(T_i | D_{A}) \times P(T_i | D_{B})$. 
    This representation is more semantic in nature as it can capture overlap in topics.
    
    
    \item Alternatively, one can take the \textit{5W1H} approach
    \citep{xie2008event}, where a given narrative $D$ can be represented in terms of unordered sets of six facets: \textit{5Ws} (Who, What, When, Where and Why) and \textit{1H} (How). In this case, we can define \textit{Semantic Overlap} as the common elements between the corresponding sets related to these 6 facets present in both narratives, i.e. $D_{ovlp} = \{S_i \}$ where $S_i$ is a set belonging to one of the six \textit{5W1H} facets. 
    It is entirely possible that one of these $S_i$'s is an empty set ($\phi$). 
    The most challenging aspect with this approach is accurately inferring the 5W1H facets.

    
    \item Another way could be to define a given document as a graph. 
    Specifically, we can consider a document $D$ as a directed graph $G=(V, E)$ where $V$ represents the vertices and $E$ represents the edges. 
    Thus, \textit{TextOverlap} can be defined as the set of common vertices or edges or both.
    Specifically, $D_{ovlp}$ can be defined as a maximum common subgraph  of both $G_A$ and $G_B$, where $G_A$ and $G_B$ are the corresponding graphs for the documents $D_A$ and $D_B$ respectively. 
    However, coming up with a graph structure $G$ which can align with both documents $D_A$ and $D_B$, would itself be a challenge. 
    
    
    \item One can also define \textit{TextOverlap} operator ($\cap$) between two documents based on historical context and prior knowledge. 
    Given a knowledge base $K$, $D_{ovlp} = \cap(D_A, D_B | K)$ \citep{radev2000common}. 
\end{enumerate}
All the approaches defined above have their specific use-cases and challenges, however, from a human-centered point of view, they may not reflect how humans generate semantic overlaps. 
A human would mostly express it in the form of natural language and this is why, we frame the \textit{TextOverlap} operator as a constraint summarization problem such that the information of the output summary is present in both the input documents. 

\section{Threshold Algorithm}
\label{app_sec:threshold_algorithm}
\vspace{-2mm}

\begin{algorithm}[!htbp] 
\caption{Threshold Function}
\label{algo:threshold}

\begin{algorithmic}[1]
    \Procedure{Threshold}{$rawSs, T$}
        \State{\textbf{initialize} $Labels \gets [ ]$}
        
        \For{each element $e$ in $rawSs$}
            \If{$e \geq t_u \% $}
                \State $Labels.append(P)$
            \ElsIf{$t_l \% \leq e \leq t_u \% $}
                \State $Labels.append(PP)$
            \Else
                \State $Labels.append(A)$
            \EndIf
        
        \EndFor
        
        \State{\textbf{return} $Labels$}
    \EndProcedure
    \end{algorithmic}

\end{algorithm}

\section{ROUGE Scores}

\begin{table}[!htb]\footnotesize 
\centering
\vspace{-2mm}
\begin{tabular}{cccc}
\toprule
\textbf{Model} & \textbf{R1} & \textbf{R2} & \textbf{RL} \\
\midrule
BART & 40.73 & 25.97 
& 29.95 \\
T5 & 38.50 & 24.63 & 27.73 \\
Pegasus & 46.36 & 29.12 & 37.41 \\
\bottomrule
\end{tabular}
\vspace{-2mm}
\caption{Average ROUGE-F1 Scores for all the test models across test dataset. For a particular sample, we take the maximum value out of the $4$ F1 scores corresponding to the 4 reference summaries. }
\vspace{-2mm}
\label{tab:rouge_score}
\vspace{-3mm}
\end{table}


\pagebreak

\begin{table*}[!t]\footnotesize
\begin{subtable}[h]{\textwidth}
\centering
\ra{1.1}
\adjustbox{max width=\textwidth}{%
\begin{tabular}{p{1.3cm}cccccccc}
\toprule
\multicolumn{9}{c}{\textbf{Machine-Human Agreement in terms of Reward Function}}  \\
\cmidrule{3-9}
\multicolumn{1}{l}{} & \multicolumn{1}{l}{} & \multicolumn{1}{c}{$\mathbf{T=(25, 75)}$} & \multicolumn{1}{c}{$\mathbf{T=(35, 65)}$} & \multicolumn{1}{c}{$\mathbf{T=(45, 75)}$}& \multicolumn{1}{c}{$\mathbf{T=(55, 65)}$} & \multicolumn{1}{c}{$\mathbf{T=(55, 75)}$} & \multicolumn{1}{c}{$\mathbf{T=(55, 80)}$} & \multicolumn{1}{c}{$\mathbf{T=(60, 80)}$}\\ 
\midrule
\multicolumn{9}{c}{\textbf{\textit{Sentence Embedding: P-v1}}}  \\
\midrule

\multirow{3}{*}{\parbox{1.3 cm}{\textbf{Precision Reward}}}
& L\textsubscript{1} &
$\SI{0.73 \pm 0.27}{}$ & $\SI{0.81 \pm 0.25}{}$ & 
$\SI{0.77 \pm 0.26}{}$ & $\SI{0.85 \pm 0.23}{}$ & 
$\SI{0.8 \pm 0.24}{}$ & $\SI{0.77 \pm 0.24}{}$ & 
$\SI{0.77 \pm 0.26}{}$ \\ 

 & L\textsubscript{2} &
$\SI{0.72 \pm 0.3}{}$ & $\SI{0.73 \pm 0.29}{}$ & 
$\SI{0.73 \pm 0.3}{}$ & $\SI{0.78 \pm 0.27}{}$ & 
$\SI{0.79 \pm 0.27}{}$ & $\SI{0.75 \pm 0.26}{}$ & 
$\SI{0.73 \pm 0.29}{}$ \\ 

 & L\textsubscript{3} &
$\SI{0.81 \pm 0.23}{}$ & $\SI{0.86 \pm 0.21}{}$ & 
$\SI{0.79 \pm 0.24}{}$ & $\SI{0.78 \pm 0.28}{}$ & 
$\SI{0.74 \pm 0.28}{}$ & $\SI{0.69 \pm 0.28}{}$ & 
$\SI{0.69 \pm 0.27}{}$ \\ 
\midrule
\multirow{3}{*}{\parbox{1.3 cm}{\textbf{Recall Reward}}} 
& L\textsubscript{1} &
$\SI{0.66 \pm 0.19}{}$ & $\SI{0.79 \pm 0.16}{}$ & 
$\SI{0.75 \pm 0.16}{}$ & $\SI{0.76 \pm 0.18}{}$ & 
$\SI{0.71 \pm 0.17}{}$ & $\SI{0.66 \pm 0.17}{}$ & 
$\SI{0.61 \pm 0.18}{}$ \\ 

 & L\textsubscript{2} &
$\SI{0.67 \pm 0.19}{}$ & $\SI{0.78 \pm 0.16}{}$ & 
$\SI{0.76 \pm 0.15}{}$ & $\SI{0.73 \pm 0.19}{}$ & 
$\SI{0.72 \pm 0.18}{}$ & $\SI{0.7 \pm 0.18}{}$ & 
$\SI{0.65 \pm 0.21}{}$ \\ 

 & L\textsubscript{3} &
$\SI{0.66 \pm 0.15}{}$ & $\SI{0.72 \pm 0.17}{}$ & 
$\SI{0.68 \pm 0.17}{}$ & $\SI{0.68 \pm 0.22}{}$ & 
$\SI{0.64 \pm 0.2}{}$ & $\SI{0.59 \pm 0.19}{}$ & 
$\SI{0.57 \pm 0.2}{}$ \\ 

\midrule
\multicolumn{9}{c}{\textbf{\textit{Sentence Embedding: STSB}}}  \\
\midrule
\multirow{3}{*}{\parbox{1.3 cm}{\textbf{Precision Reward}}}
& L\textsubscript{1} &
$\SI{0.75 \pm 0.29}{}$ & $\SI{0.75 \pm 0.29}{}$ & 
$\SI{0.75 \pm 0.29}{}$ & $\SI{0.75 \pm 0.29}{}$ & 
$\SI{0.75 \pm 0.29}{}$ & $\SI{0.75 \pm 0.3}{}$ & 
$\SI{0.75 \pm 0.23}{}$ \\ 

 & L\textsubscript{2} &
$\SI{0.63 \pm 0.32}{}$ & $\SI{0.63 \pm 0.31}{}$ & 
$\SI{0.63 \pm 0.32}{}$ & $\SI{0.63 \pm 0.31}{}$ & 
$\SI{0.63 \pm 0.32}{}$ & $\SI{0.64 \pm 0.32}{}$ & 
$\SI{0.64 \pm 0.32}{}$ \\ 

 & L\textsubscript{3} &
$\SI{0.81 \pm 0.23}{}$ & $\SI{0.82 \pm 0.23}{}$ & 
$\SI{0.81 \pm 0.23}{}$ & $\SI{0.82 \pm 0.23}{}$ & 
$\SI{0.81 \pm 0.23}{}$ & $\SI{0.81 \pm 0.22}{}$ & 
$\SI{0.81 \pm 0.22}{}$ \\ 
\midrule
\multirow{3}{*}{\parbox{1.3 cm}{\textbf{Recall Reward}}} 
& L\textsubscript{1} &
$\SI{0.66 \pm 0.21}{}$ & $\SI{0.67 \pm 0.21}{}$ & 
$\SI{0.66 \pm 0.21}{}$ & $\SI{0.68 \pm 0.21}{}$ & 
$\SI{0.67 \pm 0.21}{}$ & $\SI{0.65 \pm 0.21}{}$ & 
$\SI{0.66 \pm 0.21}{}$ \\ 

 & L\textsubscript{2} &
$\SI{0.57 \pm 0.2}{}$ & $\SI{0.58 \pm 0.21}{}$ & 
$\SI{0.57 \pm 0.2}{}$ & $\SI{0.59 \pm 0.2}{}$ & 
$\SI{0.59 \pm 0.2}{}$ & $\SI{0.58 \pm 0.2}{}$ & 
$\SI{0.58 \pm 0.21}{}$ \\ 

 & L\textsubscript{3} &
$\SI{0.67 \pm 0.19}{}$ & $\SI{0.67 \pm 0.2}{}$ & 
$\SI{0.67 \pm 0.19}{}$ & $\SI{0.68 \pm 0.2}{}$ & 
$\SI{0.68 \pm 0.19}{}$ & $\SI{0.67 \pm 0.18}{}$ & 
$\SI{0.68 \pm 0.18}{}$ \\ 
\midrule
\multicolumn{9}{c}{\textbf{\textit{Sentence Embedding: USE}}}  \\
\midrule

\multirow{3}{*}{\parbox{1.3 cm}{\textbf{Precision Reward}}}
& L\textsubscript{1} &
$\SI{0.76 \pm 0.29}{}$ & $\SI{0.77 \pm 0.3}{}$ & 
$\SI{0.78 \pm 0.27}{}$ & $\SI{0.8 \pm 0.28}{}$ & 
$\SI{0.8 \pm 0.27}{}$ & $\SI{0.77 \pm 0.27}{}$ & 
$\SI{0.8 \pm 0.27}{}$ \\ 

 & L\textsubscript{2} &
$\SI{0.69 \pm 0.32}{}$ & $\SI{0.66 \pm 0.32}{}$ & 
$\SI{0.71 \pm 0.3}{}$ & $\SI{0.68 \pm 0.3}{}$ & 
$\SI{0.72 \pm 0.3}{}$ & $\SI{0.76 \pm 0.29}{}$ & 
$\SI{0.78 \pm 0.29}{}$ \\ 

 & L\textsubscript{3} &
$\SI{0.82 \pm 0.24}{}$ & $\SI{0.85 \pm 0.22}{}$ & 
$\SI{0.85 \pm 0.23}{}$ & $\SI{0.86 \pm 0.21}{}$ & 
$\SI{0.85 \pm 0.23}{}$ & $\SI{0.82 \pm 0.23}{}$ & 
$\SI{0.78 \pm 0.25}{}$ \\ 
\midrule
\multirow{3}{*}{\parbox{1.3 cm}{\textbf{Recall Reward}}} 
& L\textsubscript{1} &
$\SI{0.64 \pm 0.19}{}$ & $\SI{0.67 \pm 0.19}{}$ & 
$\SI{0.68 \pm 0.19}{}$ & $\SI{0.7 \pm 0.21}{}$ & 
$\SI{0.69 \pm 0.22}{}$ & $\SI{0.64 \pm 0.2}{}$ & 
$\SI{0.65 \pm 0.21}{}$ \\ 

 & L\textsubscript{2} &
$\SI{0.62 \pm 0.19}{}$ & $\SI{0.63 \pm 0.2}{}$ & 
$\SI{0.66 \pm 0.18}{}$ & $\SI{0.66 \pm 0.21}{}$ & 
$\SI{0.68 \pm 0.2}{}$ & $\SI{0.68 \pm 0.19}{}$ & 
$\SI{0.69 \pm 0.21}{}$ \\ 

 & L\textsubscript{3} &
$\SI{0.64 \pm 0.16}{}$ & $\SI{0.68 \pm 0.19}{}$ & 
$\SI{0.66 \pm 0.16}{}$ & $\SI{0.69 \pm 0.2}{}$ & 
$\SI{0.65 \pm 0.19}{}$ & $\SI{0.6 \pm 0.17}{}$ & 
$\SI{0.6 \pm 0.18}{}$ \\ 
\bottomrule

\end{tabular}
}
\caption{Average Precision and Recall reward/correlation (mean $\pm$ std) between label-annotators (L\textsubscript{i}) and automatically inferred labels using SEM-F1. The results are shown for different embedding models (\ref{subsec:sem_f1_evaluation}) and multiple threshold levels $T=(t_l, t_u)$. For all the annotators L\textsubscript{i} ($i \in \{1, 2, 3\}$), correlation numbers are quite high ($\geq 0.50$).
Moreover, the reward values are consistent/stable across all the $5$ embedding models and threshold values.}
\label{tab:average_rewards_score}
\end{subtable}
\hfill
\\
\\
\begin{subtable}[h]{\textwidth}
\centering
\ra{1.1}
\adjustbox{max width=\textwidth}{%
\begin{tabular}{p{1.3cm}cccccccc}

\toprule
\multicolumn{9}{c}{\textbf{Machine-Human Agreement in terms of Kendall Rank Correlation}}  \\
\cmidrule{3-9}
\multicolumn{1}{l}{} & \multicolumn{1}{l}{} & \multicolumn{1}{c}{$\mathbf{T=(25, 75)}$} & \multicolumn{1}{c}{$\mathbf{T=(35, 65)}$} & \multicolumn{1}{c}{$\mathbf{T=(45, 75)}$}& \multicolumn{1}{c}{$\mathbf{T=(55, 65)}$} & \multicolumn{1}{c}{$\mathbf{T=(55, 75)}$} & \multicolumn{1}{c}{$\mathbf{T=(55, 80)}$} & \multicolumn{1}{c}{$\mathbf{T=(60, 80)}$}\\
\midrule
\multicolumn{9}{c}{\textbf{\textit{Sentence Embedding: P-v1}}}  \\
\midrule
\multirow{3}{*}{\parbox{1.3 cm}{\textbf{Precision Reward}}}
& L\textsubscript{1} &
$0.55$ & $0.6$ & $0.58$ & $0.59$ & $0.57$ & $0.56$ & $0.54$ \\ 

 & L\textsubscript{2} &
$0.61$ & $0.67$ & $0.63$ & $0.67$ & $0.64$ & $0.67$ & $0.68$ \\

 & L\textsubscript{3} &
 $0.54$ & $0.62$ & $0.56$ & $0.64$ & $0.6$ & $0.56$ & $0.52$ \\
\midrule
\multirow{3}{*}{\parbox{1.3 cm}{\textbf{Recall Reward}}} 
& L\textsubscript{1} &
$0.53$ & $0.64$ & $0.66$ & $0.62$ & $0.61$ & $0.62$ & $0.59$ \\

 & L\textsubscript{2} &
$0.55$ & $0.64$ & $0.67$ & $0.63$ & $0.63$ & $0.64$ & $0.61$ \\

 & L\textsubscript{3} &
$0.54$ & $0.65$ & $0.64$ & $0.66$ & $0.65$ & $0.65$ & $0.61$ \\
\midrule
\multicolumn{9}{c}{\textbf{\textit{Sentence Embedding: STSB}}}  \\
\midrule

\multirow{3}{*}{\parbox{1.3 cm}{\textbf{Precision Reward}}}
& L\textsubscript{1} &
$0.57$ & $0.67$ & $0.58$ & $0.66$ & $0.6$ & $0.57$ & $0.58$ \\

 & L\textsubscript{2} &
$0.66$ & $0.63$ & $0.65$ & $0.63$ & $0.7$ & $0.63$ & $0.6$ \\

 & L\textsubscript{3} &
$0.56$ & $0.57$ & $0.58$ & $0.56$ & $0.59$ & $0.57$ & $0.56$ \\
\midrule
\multirow{3}{*}{\parbox{1.3 cm}{\textbf{Recall Reward}}} 
& L\textsubscript{1} &
$0.55$ & $0.65$ & $0.64$ & $0.62$ & $0.62$ & $0.61$ & $0.59$ \\

 & L\textsubscript{2} &
$0.56$ & $0.65$ & $0.65$ & $0.63$ & $0.63$ & $0.64$ & $0.63$ \\

 & L\textsubscript{3} &
$0.54$ & $0.59$ & $0.61$ & $0.57$ & $0.58$ & $0.57$ & $0.54$ \\
\midrule
\multicolumn{9}{c}{\textbf{\textit{Sentence Embedding: USE}}}  \\
\midrule

\multirow{3}{*}{\parbox{1.3 cm}{\textbf{Precision Reward}}}
& L\textsubscript{1} &
$0.58$ & $0.62$ & $0.6$ & $0.61$ & $0.59$ & $0.62$ & $0.65$ \\

 & L\textsubscript{2} &
$0.68$ & $0.7$ & $0.68$ & $0.68$ & $0.68$ & $0.7$ & $0.73$ \\

 & L\textsubscript{3} &
$0.66$ & $0.67$ & $0.65$ & $0.64$ & $0.63$ & $0.53$ & $0.56$ \\
\midrule
\multirow{3}{*}{\parbox{1.3 cm}{\textbf{Recall Reward}}} 
& L\textsubscript{1} &
$0.53$ & $0.59$ & $0.56$ & $0.61$ & $0.62$ & $0.61$ & $0.6$ \\

 & L\textsubscript{2} &
$0.54$ & $0.6$ & $0.61$ & $0.62$ & $0.64$ & $0.64$ & $0.62$ \\

 & L\textsubscript{3} &
$0.52$ & $0.6$ & $0.58$ & $0.61$ & $0.61$ & $0.6$ & $0.6$ \\
\bottomrule

\end{tabular}
}
\caption{Average Precision and Recall Kendall Tau between label-annotators (L\textsubscript{i}) and automatically inferred labels using SEM-F1. The results are shown for different embedding models (\ref{subsec:sem_f1_evaluation}) and multiple threshold levels $T=(t_l, t_u)$. For all the annotators L\textsubscript{i} ($i \in \{1, 2, 3\}$), correlation numbers are quite high ($\geq 0.50$).
Moreover, the reward values are consistent/stable across all the $5$ embedding models and threshold values.
All values are statistically significant at p-value<0.05.  
}
\label{tab:average_kendall_scores}
\end{subtable}
\caption{Machine-Human Agreement}
\label{tab:raw_average_rewards}
\end{table*}

\begin{table*}[!htb]
\centering
\begin{adjustbox}{width=\linewidth,center}
\begin{tabular}{ccccc}
\toprule
\multicolumn{5}{c}{AllSides Dataset: Statistics} \\
\midrule
\multicolumn{1}{c}{Split} & 
\multicolumn{1}{c}{\#words (docs)} & 
\multicolumn{1}{c}{\#sents (docs)} & 
\multicolumn{1}{c}{\#words (reference/s)} & 
\multicolumn{1}{c}{\#sents (reference/s)} \\
\midrule
Train & $1613.69$ & $66.70$ & $67.30$ & $2.82$ \\
Test & $959.80$ & $44.73$ & $65.46/38.06/21.72/32.82$ & $3.65/2.15/1.39/1.52$ \\
\bottomrule
\end{tabular}
\end{adjustbox}
\vspace{-2mm}
\caption{
Two input documents are concatenated to compute the statistics. Four numbers for reference (\#words/\#sents) in Test split corresponds to the 4 reference intersections.
Our test dataset contains of 137 samples, wherein each sample has 4 ground truth references. Out of these 4 references, 3 of them were manually written by 3 references annotators. Thus, we generated 3*137 = 411, references in total. 
One of the recent papers, titled \citep{fabbri2021summeval}, also incorporated human annotations for only 100 samples. Following them, we created reference summaries for 150 samples which later got filtered to 137 samples due to minimum 15 words criterion as described in section \ref{sec:allsides_dataset}.
Overall, we agree that having more samples in the test dataset would definitely help a lot. But this is both time and money consuming process. We are working towards it and would like to increase the number of test samples in future. 
}
\label{tab:dataset_statistics}

\end{table*}

\section{Motivation and Applications}
\label{sec:motivation}

Multiple alternative narratives are frequent in a variety of domains, including education, health sector, and privacy, and and technical areas such as Information Retrieval/Search Engines, QA, Translation etc. In general, MNSO/TextIntersect operation can be highly effective in digesting such multi-narratives (from various perspectives) at scale and speed. Here are a few examples of use-cases.

\noindent \textbf{\underline{Peer-Reviewing}:} \emph{TextIntersect} can extract sections of multiple peer-reviews for an article that agree with one other, which can assist creating a meta-review fast.

\noindent \textbf{\underline{Security and Privacy:}} By mining overlapping clauses from various privacy policies, the \emph{TextIntersect} operation may assist real-world consumers swiftly undertake a comparative study of different privacy policies and thus, allowing them to make informed judgments when selecting between multiple alternative web-services.

\noindent \textbf{\underline{Health Sector}:}  \textit{TextIntersect} can be applied to compare clinical notes in patient records to reveal changes in a patient's condition or perform comparative analysis of patients with the same diagnosis/treatment. For example, \textit{TextIntersect} can be applied to the clinical notes of two different patients who went through the same treatments to assess the effectiveness of the treatment.

\noindent \textbf{\underline{Military Intelligence:}} If $A$ and $B$ are two intelligence reports related to a mission coming from two human agents, the \emph{TextIntersect} operation can help verify the claims in each report w.r.t. the other, thus, \textit{TextIntersect} can be used as an automated claim-verification tool.

\noindent \textbf{\underline{Computational Social Science and Journalism:}} Assume that two news agencies (with different political bias) are reporting the same real-world event and their bias is somewhat reflected through the articles they write. If $A$ and $B$ are two such news articles, then the \emph{TextIntersect} operation will likely surface the facts (common information) about the event. 

Here are some of the use-cases of MNSO in various technical areas. 

\noindent \textbf{\underline{Information Retrieval/Search Engines}:} One could summarize the common information in the multiple results fetched by a search engine for a given query and show it in separate box to the user. 
This would immensely help the to quickly parse the information rather than going through each individual article. 
If they desire, they could further explore the specific articles for more details. 

\noindent \textbf{\underline{Question Answering}:} Again, one could parse the common information/answer from multiple documents pertinent to the given query/question. 
\noindent \textbf{\underline{Robust Translation}:} Suppose you have multiple translation models which translates a given document from language $A$ to language $B$. One could further apply the \textit{TextOverlap} operator on the translated documents and get a robust translation.

In general, MNSO task could be employed in any setting where we have comparative text analysis.

\clearpage
\begin{table*}[!htb]\small
\centering

\ra{1.2}
\adjustbox{max width=\textwidth}{%
\begin{tabular}{C{0.3cm} | C{3.5cm} | C{3.5cm} | C{3.5cm} | C{5.5cm}} 
\toprule
\textbf{Idx} & \multicolumn{2}{c|}{$\mathbf{D_1}$} &
\multicolumn{2}{c}{$\mathbf{D_2}$} \\
\cmidrule{1-5}
\multirow{4}{*}{$\mathbf{1}$} & 
\multicolumn{2}{C{8cm}|}{
\hlgreen{
WASHINGTON -- U.S. intelligence and law enforcement agencies have confirmed that President Donald Trump's campaign aides and associates had constant contact with Russian intelligence officials before the election, directly contradicting public statements made by top administration officials. 
}
On Jan. 15, shortly before Trump took office, Vice President Mike Pence repeatedly said on television that there were zero contacts between the campaign and Russian officials. 
$\mathbf{\dotso}$
\hlgreen{
Pence also answered "of course not" when asked a similar question that day by "Fox News Sunday" host Chris Wallace
$\mathbf{\dotso}$
Trump himself also denied these interactions
$\mathbf{\dotso}$
}
"There's nothing that would conclude me that anything different has changed with respect to that time period," Spicer said. 
$\mathbf{\dotso}$
} &
\multicolumn{2}{C{8cm}}{
\hlgreen{
President Trump said Wednesday that new reports saying his associates had contact with Russian officials during last year's campaign are "non-sense"} and accused the U.S. intelligence community of illegally leaking information to news outlets. 
\hlgreen{
"This Russian connection non-sense} is merely an attempt to cover-up the many mistakes made in Hillary Clinton's losing campaign," Mr. Trump tweeted. 
$\mathbf{\dotso}$
Among those supposedly communicating with Russian nationals was former Trump campaign chairman Paul Manafort, the report said. 
\hlgreen{Mr. Manafort denied that he ever knowingly talked to any intelligence official "or anyone}
} \\
\cmidrule{2-5}
& \multicolumn{4}{c}{\textbf{Overlap}} \\
& \multicolumn{1}{c}{\textbf{A\textsubscript{1}}} &
  \multicolumn{1}{c}{\textbf{A\textsubscript{2}}} &
  \multicolumn{1}{c}{\textbf{A\textsubscript{3}}} &
  \multicolumn{1}{c}{\textbf{AllSides}}
  \\

\cmidrule{2-5}
&
President Trump and the Trump administration deny allegations that advisers close to Trump were in constant communication during the campaign with Russians known to US intelligence.
&
Trump denied climas that advisers close to him were in "constant communication during the campaign with Russians known to US intelligence.
&
Donald Trump and his group claimed that there is no contact with Russian officials during his last year's campaign. 
&
Russian intelligence officials made repeated contact with members of President Trump's campaign staff, according to new reports that cite anonymous U.S. officials. American agencies were concerned about the contacts but haven't seen proof of collusion between the campaign and the Russian security apparatus.
\\
\midrule
\midrule
 & \multicolumn{2}{c|}{$\mathbf{D_1}$} &
\multicolumn{2}{c}{$\mathbf{D_2}$} \\
\cmidrule{1-5}
\multirow{4}{*}{$\mathbf{2}$} & 
\multicolumn{2}{C{8cm}|}{
John McCain is out of McConnell's clutches for a week or two. 
\hlgreen{
While Sen. John McCain remains in Arizona recovering from Friday's craniotomy, surgery to remove a 5 cm blood clot from above his left eye}, business will not go on as usual in Washington. 
\hlgreen{
Majority Leader Mitch McConnell, who has to have every Republican senator voting to have a prayer of passing Trumpcare, has postponed the vote for the week or two (more likely two) that McCain's recovery will take.
}
That means there's more time for opponents to fight this thing, from the side of all of us trying to keep 22 million people from losing insurance and from the other side. 
$\mathbf{\dotso}$
\hlgreen{With both Paul and Sen. Susan Collins (R-ME) solid "no" votes on the bill}, opponents only need one more out of the eight or so who've expressed reservations about the bill and the secretive, exclusive process McConnell
} &
\multicolumn{2}{C{8cm}}{
\hlgreen{WASHINGTON - The Republican effort to repeal and replace Obamacare faces a major setback as Sen. John McCain, R-Ariz., left the nation's capital for surgery on his eye. 
Over the weekend, Senate Majority Leader Mitch McConnell, R-Ky., announced the scheduled Better Care Act vote would be delayed indefinitely because of McCain's absence.}
Subsequently, the Congressional Budget Office (CBO) also delayed its analysis of the bill. 
\hlgreen{With two Republican senators opposed to the measure, McConnell needs as least 50 "yes" votes to pass it.}
\hlgreen{Sen. Rand Paul, R-Ky., says the bill, which keeps taxes on investments and other pieces of Obamacare, doesn't go far enough. 
Moderate Sen. Susan Collins, R-Maine, is also withholding her support because it would slow the rate of growth in spending on Medicaid.}
$\mathbf{\dotso}$
} \\
\cmidrule{2-5}
& \multicolumn{4}{c}{\textbf{Overlap}} \\
& \multicolumn{1}{c}{\textbf{A\textsubscript{1}}} &
  \multicolumn{1}{c}{\textbf{A\textsubscript{2}}} &
  \multicolumn{1}{c}{\textbf{A\textsubscript{3}}} &
  \multicolumn{1}{c}{\textbf{AllSides}}
  \\

\cmidrule{2-5}
&
Sen. John McCain remains in Arizona recovering from eye surgery. Senate Majority Leader Mitch  McConnell postponed the vote due to McCain's absence. Two Republican senators opposed to the bill. Possibility of bill failing.
&
Sen. John McCain remains unavailable because of the surgery on his eye. Senate Majority Leader Mitch McConnell delayed the vote in his absence. Sen. Rand Paul and Sen. Susan Collins said "no" votes on the bill.
&
Senate Majority Leader Mitch McConnell, R-Ky., announced the scheduled health care vote would be delayed indefinitely because of McCain's absence.
&
Senate Majority Leader Mitch McConnell, R-Ky., announced the scheduled Better Care Act vote would be delayed indefinitely because of McCain's absence.
\\
\bottomrule
\end{tabular}
}
\caption{Some examples of \textit{TextOverlap} from 3 human annotators (\textbf{A\textsubscript{i}}) and the AllSides ``theme-description'' for a given document pair $\{D_1, D_2\}$. 
($\mathbf{\dotso}$) denotes some the sentences which for not shown for brevity.
More examples can be found in supplementary folder. 
As we notice, AllSides ``theme-description'' is only a proxy overlap summary of the input document pairs. 
Thus, having human annotators becomes critical but it a laborious and time-consuming part on humans end. Thus, lack of available dataset is a huge challenge for MNSO task. 
}
\label{tab:qualitative_analysis}
\vspace{-4mm}
\end{table*}

\end{document}